\documentclass[letterpaper,10pt,conference]{ieeeconf}
\IEEEoverridecommandlockouts
\usepackage{cite}

\usepackage{amsmath,amssymb,amsfonts}
\DeclareMathSizes{7.5}{7.5}{5}{5}
\usepackage{graphicx}
\usepackage{xcolor}
\usepackage{booktabs}
\usepackage{multirow}
\usepackage{arydshln}
\usepackage{url}
\usepackage{float}
\usepackage{wrapfig}
\usepackage{placeins}

\usepackage[hidelinks]{hyperref}

\graphicspath{{figures/}}

\newcommand{\method}{TraceFlow}
\newcommand{\tracebank}{TraceBank}
\newcommand{\simgate}{Similarity Gate}

\newcommand{\configurebibauthors}{\bstctlcite{BSTcontrol}}
\title{\LARGE\bfseries TraceFlow: Guiding Frozen Flow-Matching Robot Policies\newline with Success and Failure Traces}

\newif\ifanonymous
\anonymousfalse
\ifanonymous
\author{Anonymous Authors}
\else
\author{Jiaxuan Zhang$^{1,2}$, Ruizhe Liu$^{1}$, Yu Zhang$^{1}$, and Yanchao Yang$^{1,*}$%
\thanks{$^{1}$The University of Hong Kong, Hong Kong SAR, China.}%
\thanks{$^{2}$Southern University of Science and Technology, Shenzhen, China.}%
\thanks{All authors are affiliated with the HKU InfoBodied AI Lab.}%
\thanks{$^{*}$Corresponding author: Yanchao Yang.}%
}
\fi

\begin{document}
\configurebibauthors

\maketitle
\thispagestyle{empty}
\pagestyle{empty}

\begin{abstract}
A vision-language-action (VLA) policy with a flow-matching action expert
generates each action chunk (a short command sequence) by integrating a
learned velocity field; once its weights are fixed, the success or failure
of an earlier rollout cannot change the chunk generated now. Concurrent
test-time methods give a frozen policy such an input from retrieved
successes, a learned critic, a verifier, or a dynamics model, but none uses
the robot's own failed rollouts as negative evidence with nothing but a
terminal outcome bit. We introduce \method{}, a progress-aligned guidance
field that turns the action densities of retrieved successful and failed
rollouts into a bounded correction to a frozen flow-matching action expert,
using one terminal outcome bit per rollout and no other label. Its
\tracebank{} stores traces, time-ordered state--action records with a
terminal label, starts from the target-task training traces, and later
admits the deployed robot's own rollouts. On an ordered real-robot packing
task the base completes 21 of 50 trials in order, \method{} 39, and one
stacking round without any weight update 47, with wrong-sequence episodes
falling from 20 to 0. In simulation the gain is selective: with per-suite
selected settings, \method{} raises RoboMemArena Sequence from 78.92\% to
91.50\% task success and Transferring from 54.41\% to 62.00\% at stacking
round 2, leaves the 26-task aggregate unchanged, lowers Counting and
Occlusion by 1.12 and 1.42 points, and changes LIBERO-Plus (Long) by
$+1.27$ points ($p=0.0733$). Stacking gains are finite, every branch peaking
before round ten, and the bank's success-to-failure ratio predicts no
retrieval allocation.
\end{abstract}

\section{Introduction}
\label{sec:introduction}
Vision-language-action (VLA) policies pair a vision-language model (VLM)
with an action expert that generates an action chunk from Gaussian noise;
with flow matching, the expert integrates a learned velocity field over a
fixed number of solver steps \cite{brohan2023rt2,kim2024openvla,octo2024,
chi2023diffusion,black2024pi0,black2025pi05,lipman2023flow}. Once deployed
with its parameters frozen (fixed for the whole deployment), the policy has
no input through which the success or failure of an earlier rollout (one
task attempt) can change the action chunk generated now. Several concurrent
methods give a frozen policy such an input: Retrieve-then-Steer and
OptimusVLA retrieve successful experience and inject a prior into the flow
sampler \cite{zhao2026retrievethensteer,li2026optimusvla}; Guided Action Flow
and test-time Q-guidance differentiate a learned critic
\cite{yang2026guidedactionflow,zhou2026qgf}; TACO and RoboMonkey sample
several chunks and keep the one a learned verifier prefers
\cite{yang2025taco,kwok2025robomonkey}; DynaGuide steers with a learned
dynamics model \cite{du2025dynaguide}; and CCDP steers a diffusion policy
away from the failed attempts of the current episode \cite{razmjoo2025ccdp}.
None of them uses the robot's own failed rollouts, labeled by nothing but a
terminal bit, as repulsion at the current solver state next to successful
rollouts as attraction, with no learned critic, verifier, or dynamics model
and with an explicit bound on the correction; that is the gap this paper
fills.

Long-horizon manipulation makes the choice of stored trace (a time-ordered
state--action record with a terminal label) harder. An early error changes
every later observation, and states that look alike can require different
actions because task progress differs, so a trace retrieved by appearance can
match the current state in appearance and not in progress, a failure mode
reported for closed-loop retrieval
\cite{shah2026halo,li2026optimusvla,zhao2026retrievethensteer}. A successful
trace is informative only at the task phase the robot has reached, and a failed
trace only where its state and progress match the current ones, because its
terminal bit does not mark which action was wrong. Retrieval must therefore
decide which rollout, which moment within it, and how strongly its action may
move the chunk being generated.

Repeated deployment adds a second requirement. A trace store can start from the
target-task training traces and grow with the deployed robot's own successes
and failures, which avoids retraining but changes the retrieval distribution
after every round, since added traces can displace the neighbors that were
useful. The problem is therefore to turn action
outcomes labeled by one terminal bit into a correction that applies each
trace only at its matching task phase, keeps the action expert's proposal and
weights, and stays bounded as the store grows.

\noindent\textbf{We introduce \method{}, a progress-aligned guidance field that
turns the action densities of retrieved successful and failed rollouts into a
bounded correction to a frozen flow-matching action expert, using one terminal
outcome bit per rollout and no other label.}
A frozen retrieval head maps the VLM state to a retrieval token; the
\simgate{} ranks stored traces by cosine similarity and returns the top-$K_+$
successful and top-$K_-$ failed candidates; progress alignment selects the
action window that follows each matched moment; and a signed term moves the
action chunk under integration toward the retrieved successful actions and away
from the failed ones. The term is capped relative to the base velocity, and its
strength declines to zero before the final solver steps, so the base expert
alone produces the final motion. Across rounds only the \tracebank{} contents
change.

We evaluate against the same checkpoint run without guidance; the concurrent
methods above are not compared head-to-head (Sec.~\ref{sec:conclusion}). The
clearest evidence is on hardware: on an ordered three-fruit packing task the
base completes 21 of 50 trials in order with 20 wrong-sequence episodes,
\method{} 39 of 50 with 2, and one stacking round of the robot's own rollouts,
with no weight update, 47 of 50 with none (Table~\ref{tab:real-world}). In
simulation the gain is selective: with per-suite selected settings,
RoboMemArena Sequence rises from 78.92 to 91.50 task success rate (TSR, \%)
over the hierarchical PrediMem backbone \cite{lei2026robomemarena} and
Transferring from 54.41 to 62.00 at stacking round 2, while Counting and
Occlusion fall by 1.12 and 1.42 points and the 26-task aggregate is unchanged
(34.92 to 34.99 TSR); on the saturated LIBERO suites and the LIBERO-Plus
(Long) OOD variants the changes, $+1.45$ and $+1.27$ points ($p=0.0733$), are
within noise (Table~\ref{tab:capability-main}). Our contributions are:
\begin{itemize}
    \item A guidance field that turns retrieved successful and failed action
    windows into a capped, early-step correction of a frozen flow-matching
    action expert from one terminal bit per rollout, with no critic,
    verifier, dynamics model, or weight update; the cap is what keeps the
    same retrieval from collapsing the base policy (24.25 versus 57.75 TSR
    against a 56.75 base, Table~\ref{tab:rq-evidence}(a)).
    \item Evidence that the gain is specific to ordering and transfer errors:
    large on the hardware ordering task and on RoboMemArena Sequence, absent
    on the memory-dependent Counting and Occlusion suites and on the 26-task
    aggregate, with the selected and shared configurations named.
    \item A ten-round stacking study of the robot's own binary-labeled
    rollouts: gains peak within a finite number of rounds (Transferring,
    Joint admission: 54.0 to 62.0 TSR at round 2; LIBERO: 94.8 to 96.8 at
    round 7), and the best $K_+/K_-$ allocation is not predicted by the
    bank's outcome ratio (Table~\ref{tab:rq-evidence}(d)).
\end{itemize}

\section{Related Work}
\label{sec:related-work}
\textbf{Action policies.}
Generalist policies generate actions as tokens, chunks, diffusion samples,
or flow-matching solves \cite{brohan2023rt2,chi2023diffusion,black2024pi0,
black2025pi05}. \method{} adds no backbone or planner; its single change is
a guidance term inside the frozen policy's solver.

\textbf{Test-time guidance of generative policies.}
Diffusion and score models denoise iteratively, whereas flow matching and
rectified flow learn a transport field
\cite{ho2020ddpm,song2021scoresde,lipman2023flow,liu2023rectifiedflow};
classifier and classifier-free guidance change the sampling field
\cite{dhariwal2021diffusion,ho2022classifierfree}, and
\cite{feng2025guidance} gives the guided field for flow matching, the formal
basis of Eq.~\eqref{eq:traceflow-summary}. Diffuser and Diffusion-QL
guide sampling for control \cite{janner2022diffuser,wang2023diffusionql}.
For frozen robot policies, DynaGuide steers denoising with a learned dynamics
model \cite{du2025dynaguide}; Guided Action Flow and test-time Q-guidance
differentiate a learned critic trained on rollouts
\cite{yang2026guidedactionflow,zhou2026qgf}; TACO and RoboMonkey sample
chunks and keep the one a learned verifier prefers
\cite{yang2025taco,kwok2025robomonkey}; CCDP composes conditional diffusion
policies so that the failed attempts of the current episode act as negative
guidance \cite{razmjoo2025ccdp}. \method{} needs none of these learned
models: its field is a nonparametric mixture over retrieved windows of both
outcomes, its negative evidence comes from earlier episodes labeled only by
a terminal bit, and its correction is capped relative to the base velocity.

\textbf{Retrieval and memory at test time.}
Episodic control and dense dual encoders retrieve pre-encoded records by
top-$K$ similarity, and contrastive objectives organize the embedding
\cite{blundell2016episodic,pritzel2017nec,karpukhin2020dpr,lewis2020rag,
oord2018cpc,chen2020simclr,he2020moco,radford2021clip,khosla2020supcon};
the retrieval head follows these objectives. For VLAs, Retrieve-then-Steer,
the closest concurrent method, keeps progress-calibrated successful segments
in a long-term memory and injects an aggregated elite prior into an
intermediate state of the flow sampler \cite{zhao2026retrievethensteer};
OptimusVLA replaces the Gaussian initialization with a retrieved task-level
prior and adds a learned consistency module \cite{li2026optimusvla}; MAP-VLA
retrieves learned soft prompts \cite{li2025mapvla}; and a growing
demonstration pool extends a frozen policy to new tasks
\cite{park2026recap}. DynaMem and HALO reuse spatial or visuomotor history
\cite{liu2024dynamem,shah2026halo}. All reuse successes or demonstrations only. \method{} retrieves both outcomes, adds a signed score
at the current solver state instead of a prior, and stacks the deployed
policy's own binary-labeled rollouts over rounds.

\textbf{Alternatives that change weights.}
Reinforcement fine-tuning of flow VLAs improves the expert through a learned
critic \cite{wang2026qvgm}; hypernetworks, test-time training, TENT, prompt
tuning, TTT layers, and RoboTTT change generated weights or an optimized
adaptation state \cite{ha2017hypernetworks,sun2020ttt,wang2021tent,
shu2022tpt,sun2025tttlayers,jiang2026robottt}; continual learning addresses
forgetting through parameter updates and is evaluated on Continual World and
LIBERO \cite{kirkpatrick2017overcoming,lopezpaz2017gem,rolnick2019clear,
buzzega2020der,wolczyk2021continualworld,liu2023libero}. \method{} changes
no weight; its rounds change only the \tracebank{} contents, so
Sec.~\ref{sec:experiments} measures nonparametric accumulation, not
parameter learning or resistance to forgetting.

\textbf{Failure and progress supervision.}
REFLECT generates failure summaries, Robot Chain-of-Thought uses subtask
rewards, and ProcVLM synthesizes frame-level stage and progress targets
\cite{liu2023reflect,zhang2024robotcot,feng2026procvlm}; terminal binary
feedback is the sparse alternative \cite{liang2021fst}. \method{} needs only
$y\in\{0,1\}$ per trace, with no failure timestamp, stage label, severity,
or corrected segment; it is action-experience reuse, not the memory
reasoning that the Counting and Occlusion tasks of RoboMemArena test
\cite{lei2026robomemarena}.

% Queue the full-width overview on page 2 so it heads page 3.
% Keep Method text in normal flow; do not force an underfilled previous page.
\begin{figure*}[!t]
    \centering
    \includegraphics[width=\textwidth]{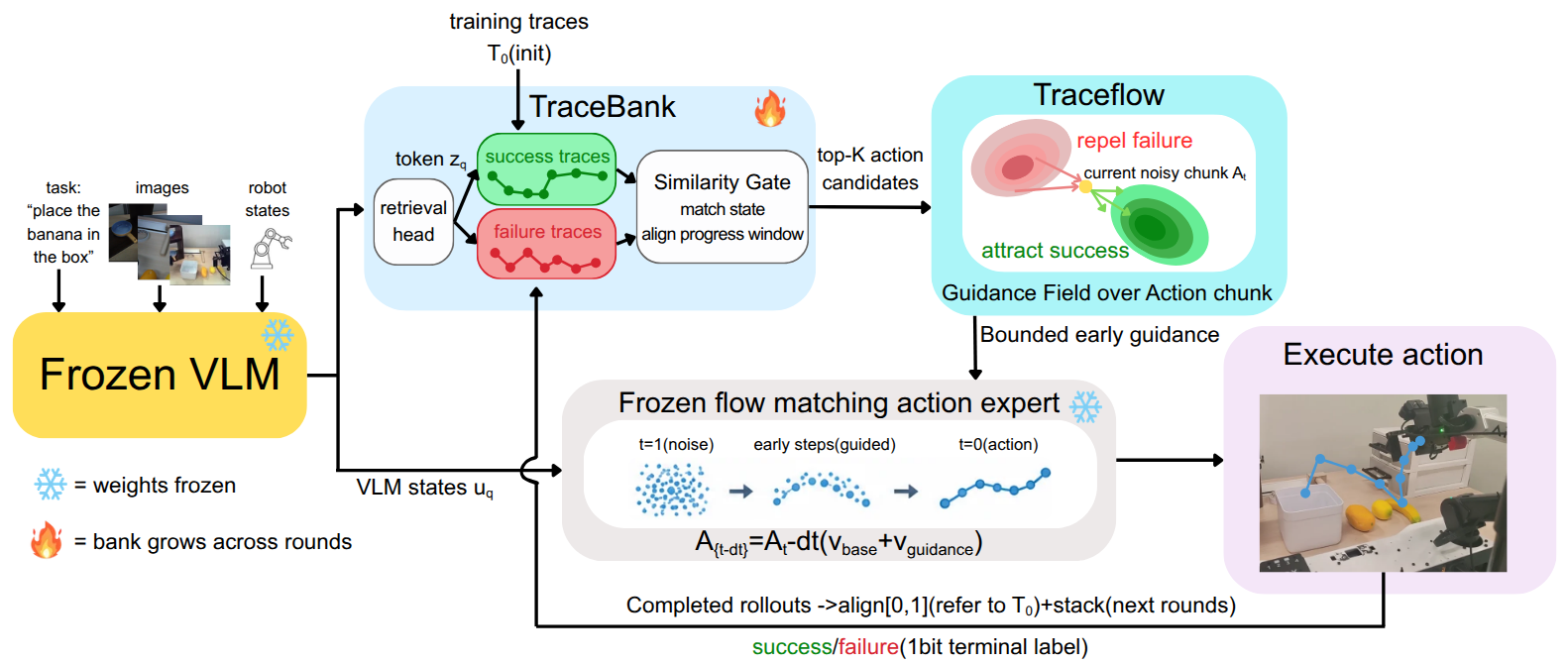}
    \caption{\method{}: observation-to-action guidance and trace reuse.
    Frozen VLM states $u_q$ condition the action expert and are encoded by a
    frozen retrieval head into query tokens $z_q$. Training traces initialize
    $\mathcal T_0$ in \tracebank{}. The \simgate{} matches states and aligns
    progress to return top-$K_+$ success and optional top-$K_-$ failure action
    windows. Their densities define a guidance field over action chunks:
    green attracts the current noisy chunk $A_t$ (yellow) toward successful
    regions, while red repels it from failed ones. Bounded guidance modifies
    early flow-integration steps before execution. Completed rollouts receive
    one terminal success/failure bit and are progress-aligned before stacking
    for subsequent rounds. Snowflakes indicate frozen weights; the flame
    indicates bank growth.}
    \label{fig:method-overview}
\end{figure*}

\section{Method}
\label{sec:method}

\subsection{Background: How a VLA Policy Produces an Action}
\label{sec:method-background}
A VLA policy maps the current camera image $o$, a language instruction $l$,
and the robot's proprioceptive state $s$ (joint or end-effector readings) to
an \emph{action chunk} $A\in\mathbb R^{H\times d}$: $H$ consecutive robot
commands, each a $d$-dimensional vector of joint or end-effector targets and
a gripper command. The policy has two parts. The \emph{VLM} is a large
transformer, pretrained on images and text, that encodes $(o,l,s)$ and an
optional history $h$ into a vector $u$ that we call the \emph{VLM state}.
The \emph{action expert} is a smaller network that turns $u$ into the chunk.
With \emph{flow matching}, the expert is a learned \emph{velocity field}
$v^{\mathrm{base}}(A,t\mid u)$: generation starts from a chunk of Gaussian
noise $A_1$ at solver time $t=1$ and moves it in $1/\Delta t$ small
\emph{Euler steps}, $A_{t-\Delta t}=A_t-\Delta t\,v^{\mathrm{base}}(A_t,t\mid u)$,
until $t=0$, where $A_0$ is the chunk the robot executes; each step costs one
evaluation of the expert (one NFE). After executing the chunk, the robot
observes again and queries the policy again, so a task is a sequence of
queries. A \emph{rollout} (or episode) is one attempt at a task from start to
success, failure, or timeout; its \emph{terminal bit} $y\in\{0,1\}$ records
the outcome. A \emph{frozen} policy is one whose weights are all fixed for
the whole deployment, the setting of this paper.

\method{} leaves this generation path unchanged and adds an external
\emph{experience path} with four modules (Fig.~\ref{fig:method-overview};
Secs.~\ref{sec:method-interface}--\ref{sec:method-stacking}): a \tracebank{}
that stores completed rollouts as traces with their terminal bit; a
retrieval step that selects stored actions comparable to the current state
and task progress; a guidance term that moves the chunk under generation
toward retrieved successful actions and away from retrieved failed ones,
with bounded strength and only in the early Euler steps; and an admission
step that appends new rollouts for later rounds. The
VLM, the action expert, and the retrieval head stay fixed throughout.

\subsection{Frozen Policy and TraceBank Interface}
\label{sec:method-interface}
A \emph{trace} is the record of one rollout: its \emph{retrieval keys}
$\mathcal Z_i$ (one vector per time step, defined below), its executed
actions $\mathcal A_i$, task and episode-boundary metadata $m_i$, and its
terminal bit $y_i$. The training rollouts of the task initialize the bank:
\begin{equation}
\label{eq:trace-record}
\tau_i=(\mathcal Z_i,\mathcal A_i,m_i,y_i),\qquad
\mathcal T_0=\{\tau_i:i\in\mathcal I_{\rm train}\},
\end{equation}
where $\mathcal I_{\rm train}$ indexes the training rollouts and
$\mathcal T_r$ denotes the bank after $r$ deployment rounds. A retrieval key
(or \emph{token}) is a fixed-length unit vector computed from a VLM state by
the \emph{retrieval head}, a small three-layer network trained offline with a
contrastive objective so that states from similar task situations receive
similar tokens; similarity between two tokens is their \emph{cosine
similarity}, the dot product of the two unit vectors (1 for identical
directions, 0 for unrelated ones). An \emph{anchor} is a stored token at one
time step of a trace together with the actions that follow it, and an
\emph{action window} is the $H$ consecutive stored actions starting at an
anchor, so that a window has the same shape as a chunk. At query $q$ the VLM
encodes the inputs, and the same state serves both paths:
\begin{equation}
\label{eq:vlm-fork}
\begin{aligned}
u_q&=\operatorname{VLM}_{\theta}(o_q,l,s_q,h_q),\\
\mathcal R_q^{\pm}&=\operatorname{Retrieve}_{\pm}(u_q,\mathcal T_r).
\end{aligned}
\end{equation}
Here $\operatorname{Retrieve}_{\pm}$ applies the retrieval head to $u_q$ to
obtain the query token $z_q$ and then the \simgate{} of Sec.~\ref{sec:method-retrieval},
returning a set $\mathcal R_q^{+}$ of successful and a set $\mathcal R_q^{-}$
of failed action windows; in parallel, $u_q$ conditions the action expert
as in the unmodified policy. Sec.~\ref{sec:experiments} lists the head
variants used for each backbone.

\subsection{Similarity Gate and Progress-Aligned Retrieval}
\label{sec:method-retrieval}
The \simgate{} computes the cosine similarity between $z_q$ and every stored
anchor and ranks anchors separately by outcome. From the success bank it
keeps the best anchor of each trace before selecting the $K_+$ highest, so
that no single trace fills the set; from the failure bank it selects the
$K_-$ highest anchors, which may come from the same trace, because a failed
rollout is negative evidence only near the states where it was recorded.
Ranking selects \emph{which rollout}; \emph{progress alignment} selects
\emph{which moment}: starting at the matched anchor, a forward-only tracker
advances along the stored trace as actions are executed and corrects its
position by local state matching, without crossing the episode boundary, so
that the returned window belongs to the task phase the robot has reached.
No progress or failure-location label is used. The returned action windows,
not their tokens, form the guidance field.

\subsection{\method{} Action Guidance}
\label{sec:method-guidance}
Retrieved windows guide, not replace, the noisy chunk
$A_t\in\mathbb R^{H\times d}$ in the policy's normalized action coordinates.
For each bank $y\in\{+,-\}$, every retrieved window $A_i^y$ receives a
weight $w_i^y=\operatorname{softmax}_i(10a_i^y)$, where $a_i^y$ is the
cosine similarity between its anchor and the query (for the failure bank
all $a_i^-$ are set to one). The windows define a \emph{kernel density}: each
window is the center of a Gaussian of width $\sigma_y$ in chunk space, and
the weighted sum of these Gaussians is a density $p_y$ whose value is high
near retrieved windows of outcome $y$. Its \emph{score}, the gradient of the
log-density with respect to the chunk, points from $A_t$ toward the nearby
windows:
\begin{equation}
\label{eq:trace-density}
\begin{aligned}
p_y(A\mid z_q)&=\sum_i w_i^y\mathcal N(A;A_i^y,\sigma_y^2I),\\
s_t^y=\nabla_{A_t}\log p_y(A_t\mid z_q)
&=\frac{\sum_i r_{i,t}^y A_i^y-A_t}{\sigma_y^2},
\end{aligned}
\end{equation}
where $r_{i,t}^y\propto w_i^y\,\mathcal N(A_t;A_i^y,\sigma_y^2I)$, normalized
over $i$. Within the retrieved set, success responsibilities combine retrieval
similarity with action proximity; failure responsibilities use uniform prior
weights and depend on action proximity. The score is therefore
recomputed as $A_t$ moves rather than fixed to one average of the windows.
Following density-based
guidance for flow matching \cite{feng2025guidance}, the success score is
added as attraction and the failure score as repulsion:
\begin{equation}
\label{eq:traceflow-summary}
\begin{aligned}
v_{q,t}^{\mathrm{guide}}&=\operatorname{Bound}\!\left(
-\lambda(t)s_t^+,\ \lambda(t)\beta_-s_t^-;
v_{q,t}^{\mathrm{base}}\right),\\
A_{t-\Delta t}&=A_t-\Delta t\left(
v_{q,t}^{\mathrm{base}}+v_{q,t}^{\mathrm{guide}}\right).
\end{aligned}
\end{equation}
Here $v_{q,t}^{\mathrm{base}}$ is the expert's own velocity at this step,
$\lambda(t)\ge 0$ is the guidance strength, $\beta_-$ weights the failure
term, and the minus sign on $s_t^+$ is what moves $A_t$ up the success
density under the backward integration of Eq.~\eqref{eq:traceflow-summary}.
$\operatorname{Bound}$ is the \emph{cap}: it rescales the success term to a
norm of at most $c_+\|v^{\mathrm{base}}\|$, the failure term to at most
$c_-\|v^{\mathrm{base}}\|$, and their sum to at most $c\|v^{\mathrm{base}}\|$,
keeping each direction unchanged; $c_+,c_-,c$ are dimensionless. Without the
cap (\emph{direct} guidance), the norm of $s_t^y$ grows with the distance
between $A_t$ and the windows and is not tied to the base velocity, so the
guidance can dominate the base flow; \emph{bounded} guidance cannot exceed a
fixed fraction of it (Table~\ref{tab:rq-evidence}(a)). A missing bank
contributes zero. The strength $\lambda(t)$ declines linearly from
$\lambda_{\max}$ at $t=1$ to zero at $t_{\mathrm{cut}}$ and stays zero for
the remaining steps, which refine contact and placement with the base expert
alone. Retrieval is done once per query; the scores are recomputed at every
active step.

\begin{figure*}[!t]
    \centering
    \includegraphics[width=0.92\textwidth]{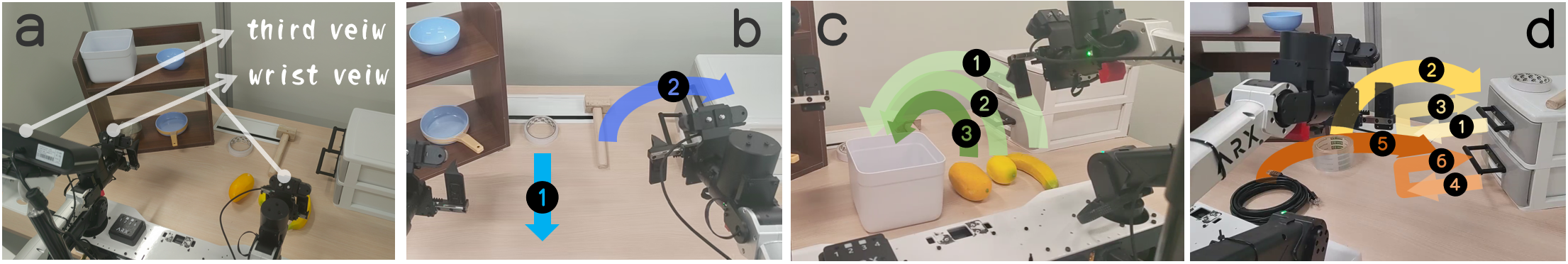}
    \caption{Real-world task suite in a cluttered storage-room scene.
    \textbf{(a)} The dual-arm platform with a third-person camera (Astra Pro
    Plus) and two wrist cameras (Gemini Pro). \textbf{(b)} T1, two-step
    relocation: move the double-sided tape down (1), then place the wooden
    hammer on the white cabinet (2). \textbf{(c)} T2, ordered packing of three
    fruits into the white box along the numbered paths. \textbf{(d)} T3, two
    ordered three-step branches across similar drawers: open the upper drawer,
    place the single-sided tape, close it (1--3); open the lower drawer, place
    the cable, close it (4--6). Arrow colors separate objects or branches; the
    numbers give the required order.}
    \label{fig:realworld-setup}
\end{figure*}

\subsection{Trace Stacking and Progress Normalization}
\label{sec:method-stacking}
After a chunk is executed, the next observation starts a new query. A
\emph{round} collects a batch of completed rollouts before the bank is
updated; \emph{stacking} appends their traces to the bank without changing
any weight. The \emph{admission rule} decides which outcomes enter:
Success-only admits new successes, Failure-only new failures, and Joint
both; existing traces stay retrievable in every case. A trace becomes
available in the next round only, so a rollout never retrieves its own
future. In this paper, continual learning means this repeated trace
accumulation, not parameter training.

Two real executions of the same task pause and move at different speeds, so
the frame index of a trace is not a measure of task progress. Before a
hardware trace $\tau$ is admitted, we express it on the normalized progress
clock $[0,1]$ of the training traces and append it to the bank of its
outcome $y_\tau$:
\begin{equation}
\label{eq:progress-summary}
\widetilde\tau=\operatorname{Align}_{[0,1]}(\tau;\mathcal T_0),\qquad
\mathcal T_{r+1}^{y}=\mathcal T_r^{y}\cup
\{\widetilde\tau\mid y_\tau=y\}.
\end{equation}
$\operatorname{Align}_{[0,1]}$ assigns each state of $\tau$ the progress
value of the training states it matches, not its elapsed time: for every
state it takes the similarity-weighted median \cite{edgeworth1888median} of
the progress values of the most similar same-task anchors (an optional
runtime estimate is clipped near this vote and blended in), and for
successful traces it projects the sequence onto a non-decreasing curve on
$[0,1]$ by pool-adjacent-violators regression \cite{best1990isotonic} and
limits each forward increment to $F_{\max}/\max(L_{\mathrm{ref}}-1,1)$,
where $L_{\mathrm{ref}}$ is the reference trace length and $F_{\max}$ the
allowed advance in reference frames. Successful runs of different lengths
thus cover the same start-to-completion range, while a failed run keeps its
local anchor match and its estimated endpoint instead of being stretched to
completion. Optional gripper-event landmarks (median normalized event times of the
training traces) keep the order of open and close events; after removing
holds, the actions are resampled on the new clock. These steps change only
the progress clock, not the outcome label or any weight.

\section{Experiments}
\label{sec:experiments}

We answer three questions on simulation and real-robot tasks. RQ1: where
does retrieved guidance help, and which integration rule keeps the base flow
intact? RQ2: does admitting binary-labeled rollouts to \tracebank{} raise
success repeatably, and for how many rounds? RQ3: does the bank's
success-to-failure ratio predict the signed retrieval allocation $K_+/K_-$?
Tables~\ref{tab:capability-main}--\ref{tab:real-world} report task success,
Fig.~\ref{fig:cl-rounds-main} the stacking rounds,
Table~\ref{tab:rq-evidence} the integration and retrieval controls, and
Table~\ref{tab:ablation-main} the ablations.

\begin{table}[!t]
\caption{\normalfont Simulation results.}
\label{tab:capability-main}
\centering
\scriptsize
\setlength{\tabcolsep}{2.1pt}
\begin{tabular}{@{}llrrr@{}}
\toprule
& & \multicolumn{2}{c}{Performance} & \\
\cmidrule(lr){3-4}
Evaluation & Backbone & Base & \textbf{\method{}} & $\Delta$ \\
\midrule
LIBERO four-suite & $\pi_{0.5}$ & 96.85 & \textbf{98.30} & +1.45 \\
LIBERO-Plus (Long) & $\pi_{0.5}$ & 79.83 & \textbf{81.10} & +1.27 \\
\midrule
Arena Seq.+Trans. & $\pi_{0.5}$ & 56.75/62.92 & \textbf{57.75/63.42} & +1.00/+0.50 \\
Arena Sequence & PrediMem & 78.92/83.09 & \textbf{91.50/95.58} & +12.58/+12.49 \\
Arena Transferring & PrediMem & 54.41/66.34 & \textbf{62.00/69.08} & +7.59/+2.74 \\
Arena Counting & PrediMem & \textbf{26.61/56.12} & 25.49/53.50 & $-$1.12/$-$2.62 \\
Arena Occlusion & PrediMem & \textbf{17.11/42.34} & 15.69/40.60 & $-$1.42/$-$1.74 \\
Arena Full26 & PrediMem & 34.92/\textbf{56.01} & \textbf{34.99}/55.04 & +0.07/$-$0.97 \\
\bottomrule
\end{tabular}
\vspace{0.5mm}

\begin{minipage}{\columnwidth}
\scriptsize
SR is success rate; TSR/CSR are task/completion success rates (\%).
Seq./Trans. abbreviate Sequence/Transferring.
PrediMem is the hierarchical backbone; Upper denotes its subtask-selector
features. Dir./Bnd. mean uncapped/bounded guidance; Joint admits both
successful and failed new traces. Bold denotes the better
value within each comparison. Base and \method{} use
the same checkpoint and evaluator within each row. These are selected
settings, not one shared configuration or exclusively pre-stacking results:
98.30 averages LIBERO suite-wise selections (one shared configuration gives
98.00); Sequence uses Upper--Dir., $K_+=50$; Transferring is Joint R2,
$8/8$. Full26 is a separate suite-conditioned aggregate, not the average
of the selected suite rows.
Counting/Occlusion rows use the suite-adapted settings of
Table~\ref{tab:ablation-main}(b). PrediMem rows use its
May checkpoint~\cite{lei2026robomemarena} with the July benchmark; the
$\pi_{0.5}$ Arena row is an eight-task, 400-episode probe. LIBERO-Plus (Long) uses
$n=2{,}519$ paired trials ($p=0.0733$).
\end{minipage}
\end{table}

\subsection{Benchmarks and Protocol}
\textbf{Simulation.} We use $\pi_{0.5}$ \cite{black2025pi05} as the reference
VLA and PrediMem \cite{lei2026robomemarena}, RoboMemArena's hierarchical
model in which an Upper VLM selects subtasks and a Lower action policy
executes them, to test whether guidance also benefits a dual-tower policy.
Both use flow-matching action experts; all policy and retrieval weights are
fixed after task-specific setup. Every comparison is against the same
checkpoint run without guidance; the concurrent test-time methods of
Sec.~\ref{sec:related-work} are not run. We evaluate on the four LIBERO suites \cite{liu2023libero}, on 2,519
paired OOD task variants of LIBERO-Plus (Long) \cite{fei2025liberoplus} (the
released \texttt{libero\_10} suite, which varies texture, viewpoint, language,
lighting, layout, initial state, and sensor noise while \tracebank{} holds
only the original LIBERO-10 traces), and on all 26 long-horizon RoboMemArena
tasks \cite{lei2026robomemarena} (Arena in the tables). LIBERO is a
saturation check. Every Arena number is re-measured in our testbed with the
official simulation and benchmark settings and the latest official
checkpoints at the time of the run, not imported from the RoboMemArena paper,
because published numbers differ across benchmark versions. TSR counts
episodes that complete every subtask in order; CSR counts stages credited
by the official task scorer, including its sequential-stage checks.
Retrieval heads: for $\pi_{0.5}$, the head takes one 2048-D VLM state (S)
or three states at frame offsets $[-20,-10,0]$ (T) through a 1024-D hidden
layer; D5/D10 sample bank anchors every five/ten frames (1024/2048-D
tokens), and T-Dense5 combines T input with D5 sampling. For PrediMem, the
head reads the subtask-selector state (Upper), the action-policy state
(Lower), or both with Upper-state age indicators (Fusion), and always
guides the Lower expert (Table~\ref{tab:ablation-main}(a1)). The Full26
comparison uses seeds 50--100, 51 trials per task, and 1,326 episodes per row;
the $\pi_{0.5}$ Arena probe uses eight tasks and 400 episodes. The reference bounded configuration uses
10 Euler steps ($\Delta t=0.1$), each with one base-velocity function
evaluation (NFE), $K_+/K_-=16/8$, $\sigma_+=\sigma_-=0.30$,
$\lambda_{\max}=0.20$, $\beta_-=0.10$, $t_{\mathrm{cut}}=0.30$,
and $(c_+,c_-,c)=(0.20,0.10,0.20)$ (Sec.~\ref{sec:method-guidance}). Changes are percentage points; paired outcomes use
exact McNemar tests only when episodes are paired.
Unless a mean is specified, each cell is one evaluation pass over the stated
trials, not a multi-seed mean. Table~\ref{tab:capability-main}'s PrediMem
Sequence/Transferring selections use 200 trials versus 204 base trials;
Full26 and Counting/Occlusion use 51 trials per task. These selected summaries
do not estimate uncertainty over configuration selection.

\textbf{Real-world.} Figure~\ref{fig:realworld-setup} shows the three cluttered-room tasks:
tape-and-hammer relocation (T1), ordered three-fruit packing (T2), and a
six-stage tape and cable task across similar drawers (T3). WS counts
wrong-sequence episodes (lower is better). Hardware TSR requires all stages
in order; hardware CSR, unlike Arena's sequential scorer, credits subtasks
irrespective of order.
We use an ARX AC-One dual-arm robot, removing its stock teleoperation fixtures
and collecting demonstrations with Meta Quest 3S controllers. The frozen
$\pi_{0.5}$ baseline receives action-expert-only fine-tuning for 30k steps at
batch size 64 with other OpenPI defaults. \method{} uses the same checkpoint,
a T-Dense5 head trained for 30 epochs (three causal anchors; five-frame
bank sampling), $K_+/K_-=8/8$, and total cap $c=0.10$; all other guidance
settings match this reference bounded configuration.

\begin{table}[!t]
\caption{\normalfont Real-world results.}
\label{tab:real-world}
\centering
\footnotesize
\renewcommand{\arraystretch}{0.95}
\setlength{\tabcolsep}{2.1pt}
\begin{tabular*}{\columnwidth}{@{\extracolsep{\fill}}llrrr@{}}
\toprule
Task & Method & TSR $\uparrow$ & WS $\downarrow$ & CSR $\uparrow$ \\
\midrule
\multirow{2}{*}{T1: tape + hammer (2)}
 & Base & 8/50 & \textbf{0/50} & 34/100 \\
 & TraceFlow & \textbf{16/50} & \textbf{0/50} & \textbf{52/100} \\
\midrule
\multirow{4}{*}{T2: ordered fruits (3)}
 & Base & 21/50 & 20/50 & 123/150 \\
 & TraceFlow & 39/50 & 2/50 & 129/150 \\
 & TraceFlow
 & \multirow{2}{*}{\textbf{47/50}}
 & \multirow{2}{*}{\textbf{0/50}}
 & \multirow{2}{*}{\textbf{145/150}} \\
 & \emph{TraceBank-Stack} & & & \\
\midrule
\multirow{2}{*}{T3: tape + cable (6)}
 & Base & 0/10 & 8/10 & 8/60 \\
 & TraceFlow & \textbf{1/10} & \textbf{0/10} & \textbf{31/60} \\
\bottomrule
\end{tabular*}
\vspace{0.5mm}

\begin{minipage}{\columnwidth}
\footnotesize
Entries count trials (TSR/WS) or annotated subtasks (CSR); bold marks the
preferable value per task. TraceBank-Stack is one additional T2 round using
joint guidance from stacked success/failure traces normalized by
Eq.~\eqref{eq:progress-summary} in Sec.~\ref{sec:method-stacking}, without a new
policy or weight update.
\end{minipage}
\end{table}

\subsection{Answering the Research Questions}

\subsubsection{RQ1: Retrieved Guidance Helps Selectively; Bounds Stabilize It}
\textbf{Real-world.} Table~\ref{tab:real-world} shows three task-specific
effects. On T1, \method{} raises TSR from 8/50 to 16/50 and CSR from 34/100
to 52/100; the base failed mostly at the tape pickup (observed, not counted
separately). On T2, the base's wrong-sequence episodes mostly swap stages 2
and 3; \method{} raises first-round TSR from 21/50 to 39/50 and lowers WS
from 20/50 to 2/50. On T3, the upper and lower drawers look alike and require
similar motions, so the base usually opens the lower drawer first and then
stalls or moves unsafely from that state; \method{} records 0/10 WS, raises
CSR from 8/60 to 31/60, and completes one of ten six-subtask episodes. With
ten T3 trials this result is descriptive.

\begin{figure}[!t]
    \centering
    \includegraphics[width=0.94\columnwidth]{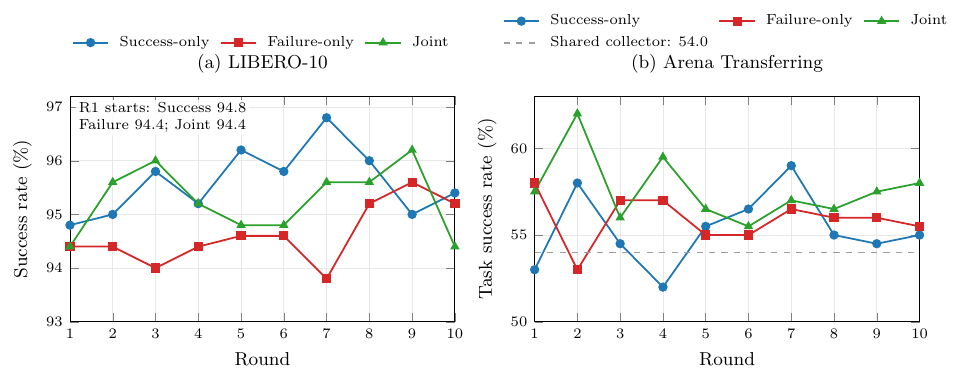}
    \caption{Ten-round trace stacking. Success-only (blue), Failure-only
    (red), and Joint (green) append new successes, failures, and both,
    respectively; previously stored traces stay queryable.
    Each round collects 500/200 rollouts in LIBERO/Arena. LIBERO
    Success-only peaks at R7 (96.8\% SR; R1: 94.8); Failure-only/Joint
    peak at R9 (95.6/96.2) and end at 95.2/94.4. Arena Success-only, Failure-only,
    and Joint peak at R7, R1, and R2 (59.0\%, 58.0\%, and 62.0\% TSR),
    respectively; Failure-only ends at 55.50/64.50 TSR/CSR.
    The dashed line is the shared Arena collector (54.0\%), the policy that
    generated the rollouts before stacking; R1 starts are LIBERO's
    empty-bank scores. R denotes the collection round.}
    \label{fig:cl-rounds-main}
\end{figure}

\textbf{Simulation.} On the eight-task $\pi_{0.5}$ probe of Sequence and
Transferring (Table~\ref{tab:rq-evidence}(a); $K_+=8$, $n=400$), the same
retrieval gives 24.25/41.62 TSR/CSR with the uncapped term and 57.75/63.42
with the cap, against a 56.75/62.92 base; the capped result is 1.00/0.50 points above
the base. The three-seed LIBERO-10 probe in Table~\ref{tab:rq-evidence}(a) gives
95.37/95.07\% direct/bounded mean SR ($p=0.580$), without a detected
paired difference. In
Table~\ref{tab:capability-main}, the gains concentrate on PrediMem Sequence
(78.92 to 91.50 TSR, Upper--Dir., $K_+=50$) and Transferring (54.41 to
62.00, Joint stacking at R2); under the shared configuration the 26-task
aggregate changes by $+0.07/-0.97$ TSR/CSR, and Counting and Occlusion fall
1.12/2.62 and 1.42/1.74 below their bases, so the simulation gain is
specific to the ordering and transfer suites. On LIBERO the four-suite
success rate rises from 96.85 to 98.30, and on LIBERO-Plus (Long) from 79.83
to 81.10 ($n=2{,}519$ paired trials, $p=0.0733$).

\subsubsection{RQ2: Binary-Labeled Stacking Improves Briefly}
\textbf{Simulation.} The three branches in Fig.~\ref{fig:cl-rounds-main}
differ only in the admission rule: Success-only appends new successes,
Failure-only new failures, and Joint both; earlier traces stay retrievable.
Every Arena branch keeps a fixed eight-task bank of 800 successful demonstrations
from four Sequence and four Transferring tasks. The LIBERO self-collection
study instead begins R1 with empty banks. Each branch is one ten-round
history with seed 7, not independently replicated histories. Admission raises
success for a finite number of rounds: LIBERO Success-only rises from 94.8\%
at R1 to its 96.8\% peak at R7, and Arena Joint from the 54.0\% collector
to 62.0\% at R2; every branch is
below its peak at R10 (Fig.~\ref{fig:cl-rounds-main}). Arena Failure-only
peaks at R1 (58.00/67.17 TSR/CSR) and ends at 55.50/64.50 in R10
(Fig.~\ref{fig:cl-rounds-main}(b)), above the collector and below Joint at
both the peak and R10. R7 and R2 are observed peaks of single histories, not stopping rules. Table~\ref{tab:rq-evidence}(d) re-evaluates the three $8/8$ branches
under $16/8$ and $8/16$ without recollecting banks.

\begin{table}[t]
\caption{\normalfont Guidance and TraceBank studies.}
\label{tab:rq-evidence}
\centering
\footnotesize
\renewcommand{\arraystretch}{0.95}
\setlength{\tabcolsep}{2pt}
\begin{tabular*}{\columnwidth}{@{\extracolsep{\fill}}lrrr@{}}
\toprule
\multicolumn{4}{@{}l}{\textbf{(a)} Integration rule, $\pi_{0.5}$ (Arena: Seq.+Trans., $K_+=8$, $n=400$)} \\
\addlinespace[1pt]
 & Base & Direct & Bounded \\
\midrule
Arena TSR/CSR & \underline{56.75/62.92} & 24.25/41.62 & \textbf{57.75/63.42} \\
L10 mean SR & 94.60 & \textbf{95.37} & \underline{95.07} \\
\midrule
\end{tabular*}

\begin{minipage}[t]{0.520\columnwidth}
\vspace{2pt}\raggedright
\textbf{(b)} Retrieval breadth $K_+$ (26 trials/task)\par
\vspace{1pt}
\setlength{\tabcolsep}{1pt}
\begin{tabular*}{\linewidth}{@{\extracolsep{\fill}}crr@{}}
$K_+$ & Seq. (Dir.) & Trans. (Bnd.) \\
\midrule
8 & 90.38/94.23 & \textbf{57.69/68.11} \\
16 & \underline{92.31/95.35} & 50.96/62.50 \\
32 & \textbf{93.27}/95.03 & 54.81/64.10 \\
50 & \textbf{93.27/95.51} & \underline{57.69/66.03} \\
100 & 89.42/92.15 & 54.81/63.14 \\
\end{tabular*}
\end{minipage}\hfill
\begin{minipage}[t]{0.460\columnwidth}
\vspace{2pt}\raggedright
\textbf{(c)} TraceBank scope (Seq.; Dir.)\par
\vspace{1pt}
\setlength{\tabcolsep}{0.5pt}
\begin{tabular*}{\linewidth}{@{\extracolsep{\fill}}llrr@{}}
Scope & Bank & $n$ & Seq. \\
\midrule
Single & Seq. & 200 & \underline{83.50}/86.58 \\
Two & Seq.+Trans. & 200 & \textbf{90.50/92.33} \\
Four & Full26 & 204 & 80.39/\underline{89.13} \\
\end{tabular*}
\end{minipage}

\vspace{2pt}
\begin{tabular*}{\columnwidth}{@{\extracolsep{\fill}}lcrr@{}}
\midrule
\multicolumn{4}{@{}l}{\textbf{(d)} Admission rule $\times$ $K_+/K_-$ allocation (Trans.; Bnd.; R1--R10)} \\
\addlinespace[1pt]
Admission & $K_+/K_-$ & Mean TSR/CSR & Peak TSR/CSR (round) \\
\midrule
Joint & $8/8$ & \textbf{57.60/65.92} & \textbf{62.00/69.08} (R2) \\
Joint & $16/8$ & 56.10/64.88 & 58.00/65.25 (R9) \\
Joint & $8/16$ & \underline{56.30/65.03} & 59.00/66.75 (R1) \\
\addlinespace[3pt]
Success & $8/8$ & 55.30/64.62 & 59.00/66.75 (R7) \\
Success & $16/8$ & \textbf{56.30/65.47} & 59.50/\underline{67.75} (R10) \\
Success & $8/16$ & \textbf{56.30}/\underline{65.04} & 59.50/66.83 (R7) \\
\addlinespace[3pt]
Failure & $8/8$ & \underline{55.90/64.85} & 58.00/67.17 (R1) \\
Failure & $16/8$ & \textbf{57.10/65.34} & \underline{60.50}/67.33 (R8) \\
Failure & $8/16$ & 55.80/64.84 & \underline{60.50}/67.67 (R9) \\
\bottomrule
\end{tabular*}
\vspace{0.5mm}

\begin{minipage}{\columnwidth}
\footnotesize
TSR/CSR (\%) except L10 (LIBERO-10) mean SR; bold/underline mark best/second-best. (b) retrieves from the success bank only.
L10 uses seeds 7/17/27 with 1,000 episodes each: Base 93.5/94.1/96.2,
Dir. 95.6/94.9/95.6, Bnd. 95.6/94.3/95.3; the pooled Dir.--Bnd.
McNemar test gives $p=0.580$.
Seq./Trans.: Sequence/Transferring; Dir./Bnd.: direct (uncapped)/bounded
($c=0.20$); Table~\ref{tab:ablation-main} shares these conventions.
Backbones: (a) $\pi_{0.5}$; (b)--(d) PrediMem. In (c), Full26 covers all four
suites, so bank content changes with bank size. In (d), means are ranked
within admission groups and peaks across all nine settings; all allocations
re-evaluate the three $8/8$ branches of Fig.~\ref{fig:cl-rounds-main}(b)
without new collection. Success/Failure abbreviate Success-only/Failure-only
admission. Mean averages R1--R10; Peak reports the highest-TSR round and
its accompanying CSR, not separately maximized metrics.
\end{minipage}
\end{table}

\textbf{Real-world.} On T2, TraceBank-Stack admits the first round's
successful and failed rollouts, normalized by Eq.~\eqref{eq:progress-summary},
and applies joint guidance with no weight update; the stacked round reaches
47/50 TSR, 145/150 CSR, and 0/50 WS (Table~\ref{tab:real-world}). Progress
normalization biases the retrieved windows toward the actions stored at the
matched phase but does not replay a trajectory: the frozen VLA is re-queried
at every observation and the guidance stays bounded. In hardware rollouts we
observed, without counting them, recoveries from local deviations back to
the intended sequence.

\subsubsection{RQ3: Sweeps Establish No Reliable Bank-Ratio Allocation Rule}
\textbf{Simulation.} Table~\ref{tab:rq-evidence}(b)--(d) varies retrieval
breadth, bank scope, and signed allocation. Sequence peaks at $K_+=32$--50,
Transferring at $K_+=8$ by CSR; both decline at $100$ (b). The two-suite bank
leads on Sequence (c), but content and size change together. Joint has its
highest mean CSR at $8/8$, Success and Failure at $16/8$; within-branch mean
TSR varies by at most 1.5 points (d). Comparing admission branches, not
measured bank ratios against optimal allocations, establishes no reliable
ratio rule without excluding such a relationship. Hardware budgets remain
empirical.

\begin{table}[!t]
\caption{\normalfont Retrieval and failure ablations.}
\label{tab:ablation-main}
\label{tab:co-oc-ablation}
\centering
\footnotesize
\renewcommand{\arraystretch}{0.95}
\setlength{\tabcolsep}{2pt}
\hrule height \heavyrulewidth
\vspace{2pt}
\begin{minipage}[t]{0.540\columnwidth}
\vspace{0pt}\raggedright
\textbf{(a1)} Head source $\times$ integration (8-task bank; $K_+/K_-=16/0$; $n=400$)\par
\vspace{1pt}
\setlength{\tabcolsep}{0.4pt}
\fontsize{7.5}{9}\selectfont
\begin{tabular*}{\linewidth}{@{\extracolsep{\fill}}lrr@{}}
Head, rule & Seq. & Trans. \\
\midrule
Lower, Dir. & 73.50/83.67 & 26.50/36.17 \\
Upper, Dir. & \textbf{90.50/92.33} & 50.50/60.17 \\
Fusion, Dir. & \underline{84.50}/90.67 & 49.00/62.58 \\
Lower, Bnd. & 77.50/84.75 & \underline{55.50/65.58} \\
Upper, Bnd. & 83.50/\underline{91.08} & 54.00/62.50 \\
Fusion, Bnd. & 78.00/87.33 & \textbf{58.00/68.50} \\
\end{tabular*}
\end{minipage}\hfill
\begin{minipage}[t]{0.440\columnwidth}
\vspace{0pt}\raggedright
\textbf{(a2)} S/T = 1/3 input anchors (8-task agg.; $K_+=8$; $n=400$)\par
\vspace{1pt}
\setlength{\tabcolsep}{0.3pt}
\fontsize{7.5}{9}\selectfont
\begin{tabular*}{\linewidth}{@{\extracolsep{\fill}}lrr@{}}
Setting & S & T \\
\midrule
D10, Dir. & \textbf{33.75/47.83} & \underline{24.25/39.25} \\
D10, Bnd. & \textbf{55.00/62.33} & \underline{54.25/61.00} \\
D5, Bnd. & \underline{55.50/63.17} & \textbf{57.75/63.42} \\
\end{tabular*}
\end{minipage}

\vspace{2pt}
\setlength{\tabcolsep}{0.3pt}
\begin{tabular*}{\columnwidth}{@{\extracolsep{\fill}}lllcrr@{}}
\midrule
\multicolumn{6}{@{}l}{\textbf{(b)} Counting/Occlusion audit: integration $\times$ bank scope (TSR/CSR)} \\
\addlinespace[1pt]
Setting & Integration & Bank & $n$/task & Counting & Occlusion \\
\midrule
Frozen base & none & none & 51 & \textbf{26.61/56.12} & \textbf{17.11/42.34} \\
Full26 & Dir. & Full26 & 51 & 15.97/37.16 & 8.38/33.69 \\
Fixed cap & Bnd. (.20) & Full26 & 51 & 23.53/48.83 & 13.90/39.50 \\
Adaptive & Bnd. (fitted) & Full26 & 10 & 21.43/49.05 & 11.82/36.10 \\
Suite adaptive & Bnd. (.01/.04) & suite-only & 51 & \underline{25.49/53.50} & \underline{15.69/40.60} \\
\end{tabular*}

\begin{tabular*}{\columnwidth}{@{\extracolsep{\fill}}llrr@{}}
\midrule
\multicolumn{4}{@{}l}{\textbf{(c)} Persistent Upper-error audit (Full26; Bnd.; guidance minus base)} \\
\addlinespace[1pt]
Suite & Metric & $\Delta$ (pp) & $p$ \\
\midrule
Counting & Pre-error progress & $-4.67$ & .049 \\
Counting & Post-error completion & $-7.47$ & .037 \\
Occlusion & Pre-error progress & $+0.15$ & .578 \\
Occlusion & Post-error completion & $-9.11$ & $9.34\times10^{-5}$ \\
\bottomrule
\end{tabular*}
\vspace{0.5mm}

\begin{minipage}{\columnwidth}
\footnotesize
Blocks (a1)/(b)/(c) use PrediMem; (a2) uses $\pi_{0.5}$.
Upper/Lower denote subtask-selector/action-policy feature sources;
Fusion combines both. The 8-task bank contains 800 successful demonstrations
from four Seq. and four Trans. tasks.
D5/D10 mean five-/ten-frame anchor sampling (1024-/2048-D output tokens).
S/T in (a2) denote single-/three-anchor inputs, not
task suites; (a1) reports Seq. and Trans. separately and (a2) their eight-task
aggregate. In (b), caps are given per row; the 10-trial Adaptive row is an
unranked matched-seed probe. Suite-only restricts the bank to the evaluated
suite; fitted denotes the adaptive cap probe, not a fixed cap. Suite
adaptation changes cap and bank scope together. Cells are single passes
over the stated trials; (a1)/(a2) use 400 episodes per setting.
Block (c) uses bounded guidance, Full26 T-Dense5 Fusion,
$K_+=16$, and 51 trials/task; the boundary is the retrospectively identified
start of the persistent wrong-Upper suffix. Pre-error progress: fraction of
all stages completed before it; post-error completion: fraction of remaining
stages completed in episode pairs with the same boundary goal rank (different
denominators, not task SR). $p$ from an exact sign test (pre) and exact
McNemar (post). A diagnostic, not causal evidence.
\end{minipage}
\end{table}

\subsection{Ablation Studies}

Table~\ref{tab:ablation-main} reports the Arena ablations. In (a1), Sequence
is best with the Upper source and direct integration and Transferring with
the Fusion source and bounded integration
, so the
source--integration preference is task-dependent. In (a2), three-anchor input
(T) lowers the Dense10 head (D10: anchors every ten frames)
but raises bounded Dense5 (D5: every five frames) from 55.50/63.17 to 57.75/63.42 TSR/CSR on the
eight-task aggregate, which is why T-Dense5 is the unified head; we claim no
general advantage of temporal context. D5/D10 also use 1024/2048-D output
tokens, respectively, so only S versus T within one density fixes both
sampling and output dimension (Sec.~\ref{sec:method-retrieval}).

In (b), the cap and the suite-restricted bank reduce the Counting and
Occlusion deficits (from 15.97/37.16 and 8.38/33.69 with the uncapped Full26
bank to 25.49/53.50 and 15.69/40.60 with suite adaptation), but no setting
reaches the base (26.61/56.12 and 17.11/42.34). Suite adaptation changes the
cap and the bank scope together, so it does not isolate the effect of a
weaker guidance. Table~\ref{tab:ablation-main}(c) examines whether the deficits
arise only after persistent Upper errors. Occlusion changes by $+0.15$ pp in
pre-error progress and $-9.11$ pp in post-error completion; Counting already
loses $4.67$ pp in pre-error progress. Thus Upper error alone does not explain
both suites; this retrospective association supplies no causal explanation or
online switch criterion. These deficits remain unresolved.

\section{Limitations and Conclusion}
\label{sec:conclusion}
Four limits bound the claims. First, the guidance moves the action chunk
toward stored actions and is not designed to reconstruct missing upstream
event information: Counting and Occlusion stay below the base under every
tested cap and bank scope, and the audit of Table~\ref{tab:ablation-main}(c)
associates part of the Occlusion loss with persistent subtask-selection
errors without establishing a cause. Second, the simulation headline cells
are per-suite selected configurations evaluated in one pass, the 26-task
aggregate under the shared configuration is unchanged, and the LIBERO and
LIBERO-Plus changes are within noise; the T3 result rests on ten trials.
Third, every admitted trace changes which anchors rank highest, so stacking
gains are finite and branch-specific (each branch peaks between R1 and R9
and is below its peak at R10, Fig.~\ref{fig:cl-rounds-main}), the ten-round
histories are single runs, and the sweeps establish no bank-ratio rule for
$K_+/K_-$ (Table~\ref{tab:rq-evidence}(d)). Fourth, the concurrent test-time
methods of Sec.~\ref{sec:related-work} are not compared head-to-head; the
only baseline is the same checkpoint without guidance.

\method{} adds a bounded, progress-aligned success and failure field to a
frozen flow-matching action expert, using one terminal bit per rollout and no
weight update. Its effect is specific: it removes ordering errors on a real
robot (21 to 39 of 50 ordered completions, 47 after one stacking round) and
raises RoboMemArena Sequence from 78.92 to 91.50 TSR, and it does not help
where a task fails for lack of event information in the policy state.

% \section*{Acknowledgments and Disclosure of Funding}
% This work was supported by X under Project X (Grant No.~X), X under Project X
% (Grant No.~X), and X under Grant No.~X. The authors thank X for providing
% experimental facilities and computational resources, and acknowledge X for
% technical assistance and helpful discussions.

\bibliographystyle{IEEEtran}
\bibliography{references}

\end{document}